# Learning Determinantal Point Processes

**Alex Kulesza**
Dept. of Computer and Information Science
University of Pennsylvania
Philadelphia, PA 19104
`kulesza@cis.upenn.edu`

**Ben Taskar**
Dept. of Computer and Information Science
University of Pennsylvania
Philadelphia, PA 19104
`taskar@cis.upenn.edu`

## Abstract

Determinantal point processes (DPPs), which arise in random matrix theory and quantum physics, are natural models for subset selection problems where diversity is preferred. Among many remarkable properties, DPPs offer tractable algorithms for exact inference, including computing marginal probabilities and sampling; however, an important open question has been how to learn a DPP from labeled training data. In this paper we propose a natural feature-based parameterization of conditional DPPs, and show how it leads to a convex and efficient learning formulation. We analyze the relationship between our model and binary Markov random fields with repulsive potentials, which are qualitatively similar but computationally intractable. Finally, we apply our approach to the task of extractive summarization, where the goal is to choose a small subset of sentences conveying the most important information from a set of documents. In this task there is a fundamental tradeoff between sentences that are highly relevant to the collection as a whole, and sentences that are diverse and not repetitive. Our parameterization allows us to naturally balance these two characteristics. We evaluate our system on data from the DUC 2003/04 multi-document summarization task, achieving state-of-the-art results.

## 1 INTRODUCTION

Many real-world tasks can be cast as subset selection problems where the goal is to choose a high-quality but diverse set of items from some base set. For example, in extractive multi-document summarization, a system chooses from among the sentences in a collection of documents a subset that summarizes the most important information. This selection problem is a balancing act: on the one hand, each selected sentence should be relevant, sharing significant information with the collection as a whole; on the other, the selected sentences should be diverse as a group so that the summary is not repetitive and is as informative as possible given its length. The modeling and learning of such problems can be challenging, as the diversity requirement implies strong repulsive interactions between variables. Standard models like Markov random fields often involve notoriously intractable inference problems when such negative correlations are present.

Determinantal point processes (DPPs), which arise in quantum physics and have been studied extensively by probabilists (Macchi, 1975; Hough et al., 2006; Borodin, 2009), are probabilistic models that give the likelihood of selecting a subset of items as the determinant of a kernel matrix. Viewed as joint distributions over the binary variables corresponding to item selection, DPPs can capture only negative correlations. This makes them significantly less general than Markov random fields. However, they are a natural fit for the problems described above; furthermore, unlike Markov random fields with negative interactions, DPPs offer remarkably tractable inference: computing marginals, computing certain conditional probabilities, and sampling can all be done exactly in polynomial time.

Kulesza and Taskar (2010) recently used DPPs to select multiple poses of people in an image (where the poses are "diverse" in that they tend not to overlap). However, the question of learning DPPs from data has not yet been addressed. In this work, we propose a conditional version of the DPP, and show how it can be parameterized to allow for convex and efficient maximum-likelihood learning. The computation of feature expectations is an important step during training, and we show how it can be done exactly and efficiently. We also describe an algorithm for approximating at test time the maximum *a posteriori* prediction. We demonstrate the utility of our model on an extractive multi-document summarization task.

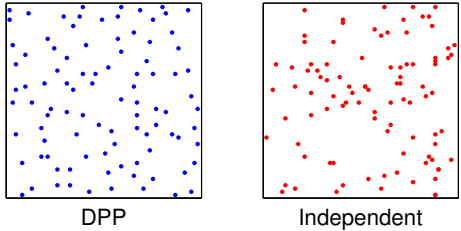

Figure 1: A set of points in the plane drawn from a DPP (left), and the same number of points sampled independently using a Poisson process (right).

The paper proceeds as follows. In Section 2, we define DPPs, give the intuition behind their unique characteristics, and show how to compute some useful quantities. In Section 3, we compare the distributions realizable by DPPs and MRFs to give a sense of their relative expressiveness. In Section 4, we describe conditional DPPs suitable to feature-based parameterization and learning. In Section 5, we derive a convex maximum-likelihood estimation approach for learning the parameters of our model, and we discuss inference in Section 6. In Section 7 we show state-of-the-art performance of the trained model on DUC 2004 extractive summarization data.

## 2 DETERMINANTAL PROCESSES

A point process $\mathcal{P}$ on a discrete set $\mathcal{Y} = \{1, \ldots, N\}$ is a probability measure on $2^{\mathcal{Y}}$, the set of all subsets of $\mathcal{Y}$. $\mathcal{P}$ is called a determinantal point process (DPP) if, when $\mathbf{Y}$ is a random set drawn according to $\mathcal{P}$, we have, for every $A \subseteq \mathcal{Y}$:

$$\mathcal{P}(A \subseteq \mathbf{Y}) = \det(K_A) , \qquad (1)$$

for some positive semidefinite matrix $K \preceq I$ (all eigenvalues of K are less then or equal to 1) indexed by the elements of $\mathcal{Y}$. $K_A \equiv [K_{ij}]_{ij \in A}$ denotes the restriction of $K$ to the entries indexed by elements of $A$, and we adopt $\det(K_\emptyset) = 1$. We will refer to $K$ as the marginal kernel, as it contains all the information needed to compute the probability of any subset $A$ being included in $\mathbf{Y}$. A few simple observations follow from Equation (1):

$$\mathcal{P}(i \in \mathbf{Y}) = K_{ii} \qquad (2)$$
$$\mathcal{P}(i, j \in \mathbf{Y}) = K_{ii} K_{jj} - K_{ij} K_{ji} \qquad (3)$$
$$= \mathcal{P}(i \in \mathbf{Y}) \mathcal{P}(j \in \mathbf{Y}) - K_{ij}^2 .$$

That is, the diagonal of $K$ gives the marginal probabilities of inclusion for individual elements of $\mathcal{Y}$, and the off-diagonal elements determine the (anti-) correlations between pairs of elements: large values of $K_{ij}$ imply that $i$ and $j$ tend not to co-occur. A DPP might therefore be used to model diverse sets of items, for example, sentences in a summary. Note that DPPs cannot represent distributions where elements are *more* likely to co-occur than if they were independent: correlations are always negative.

Figure 1 shows the difference between sampling a set of points in the plane using a DPP (with $K_{ij}$ inversely related to the distance between points $i$ and $j$), which leads to a widely spread set with good coverage, and sampling points independently, where the points exhibit random clumping. Determinantal point processes, introduced to model fermions (Macchi, 1975), also arise in studies of non-intersecting random paths, random spanning trees, and eigenvalues of random matrices (Daley and Vere-Jones, 2003; Borodin and Soshnikov, 2003; Hough et al., 2006).

For the purposes of modeling real data, however, the most relevant construction of DPPs is not through $K$ but via L-ensembles (Borodin, 2009). An **L-ensemble** defines a DPP via a positive semidefinite matrix $L$ indexed by the elements of $\mathcal{Y}$:

$$\mathcal{P}_L(\mathbf{Y} = Y) = \frac{\det(L_Y)}{\det(L + I)} , \qquad (4)$$

where $I$ is the $N \times N$ identity matrix. As a shorthand, we will write $\mathcal{P}_L(Y)$ instead of $\mathcal{P}_L(\mathbf{Y} = Y)$ when the meaning is clear. Note that $\mathcal{P}_L$ is normalized due to the identity

$$\sum_{Y \subseteq \mathcal{Y}} \det(L_Y) = \det(L + I) . \qquad (5)$$

$K$ and $L$ offer alternative representations of DPPs, and we can easily translate between the two; for example, we can compute the marginal kernel $K$ for an L-ensemble:

$$K = (L + I)^{-1} L . \qquad (6)$$

Note that $K$ can be computed from an eigendecomposition of $L = \sum_{n=1}^{N} \lambda_n \boldsymbol{v}_n \boldsymbol{v}_n^\top$ by a simple rescaling of eigenvalues:

$$K = \sum_{n=1}^{N} \frac{\lambda_n}{\lambda_n + 1} \boldsymbol{v}_n \boldsymbol{v}_n^\top . \qquad (7)$$

We can also similarly compute $L = K(I - K)^{-1}$, as long as the inverse exists.

Under both $K$ and $L$ representations, subsets that have higher diversity, as measured by the corresponding kernel, have higher likelihood. However, while $K$ gives rise to marginal probabilities, L-ensembles directly model the probabilities of atomic events, offering a convenient representation for optimization. Furthermore, $L$ need only be positive semidefinite, while the eigenvalues of $K$ are bounded above. For these reasons we focus our modeling efforts on DPPs represented as L-ensembles.

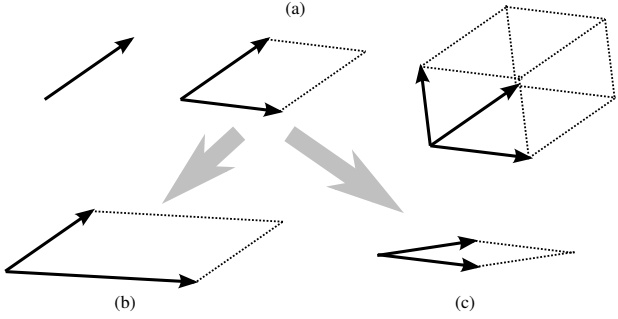

Figure 2: A geometric view of DPPs: each vector corresponds to an item. (a) The probability of a selected set is the square of the volume spanned by its included item vectors. (b) As the quality of an item increases, so do the probabilities of sets containing that item. (c) As the similarity between two items increases, the probabilities of sets containing both of them decrease.

Following Kulesza and Taskar (2010), we can decompose the kernel $L$ as a Gram matrix:

$$L_{ij} = q_i \phi_i^\top \phi_j q_j \ . \qquad (8)$$

We can think of $q_i \in \mathbb{R}^+$ as measuring the intrinsic "quality" of an item (e.g., sentence) $i$, and $\phi_i \in \mathbb{R}^n$, $\|\phi_i\|_2 = 1$ as a normalized feature vector describing an item so that $\phi_i^\top \phi_j \in [-1, 1]$ is a measure of "similarity" between items $i$ and $j$. We use the following shorthand for similarity:

$$S_{ij} \equiv \phi_i^\top \phi_j = \frac{L_{ij}}{\sqrt{L_{ii} L_{jj}}} \ .$$

The decomposition of $L$ serves two purposes. First, it implicitly enforces the constraint that $L$ must be positive semidefinite, which simplifies learning. Second, it allows us to independently model quality and similarity, and then combine them into a unified model. In particular, we have:

$$\mathcal{P}_L(\mathbf{Y} = Y) = \left( \prod_{i \in Y} q_i^2 \right) \det(S_Y) \ . \qquad (9)$$

The first term increases with the quality of the selected items, while the second term increases with the diversity of the selected items.

If we think of each item $i$ as being represented by the vector $q_i \phi_i$, then the determinant of $L_Y$ has an intuitive geometric interpretation: it is the squared volume of the parallelepiped spanned by the vectors representing the items in $Y$. Figure 2(a) illustrates this for $Y$ of cardinalities 1–3. If we increase the quality of an item $i$, then it is easy to see (Figure 2(b)) that the probability of any set containing $i$ will also increase. On the other hand, if $i$ and $j$ become more similar to one another (Figure 2(c)), the probability of any set containing the pair will decrease. Thus our model naturally balances the two objectives of high quality and high diversity.

In addition to computing marginals (Equation (1)) and the normalizing constant (Equation (5)), a surprising number of other DPP inference operations are also efficient, despite the fact that we are modeling an exponential number of possible subsets $Y$.

For example, we can compute conditional probabilities:

$$\mathcal{P}(\mathbf{Y} = A \cup B \mid A \subseteq \mathbf{Y}) = \frac{\det(L_{A \cup B})}{\det(L + I_{\mathcal{Y} \setminus A})} \ , \qquad (10)$$

where $I_{\mathcal{Y} \setminus A}$ is the matrix with ones in the diagonal entries indexed by elements of $\mathcal{Y} \setminus A$ and zeros everywhere else. Conditional marginal probabilities $\mathcal{P}(B \subseteq \mathbf{Y} \mid A \subseteq \mathbf{Y})$ as well as inclusion/exclusion probabilities $\mathcal{P}(A \subseteq \mathbf{Y} \wedge B \cap \mathbf{Y} = \emptyset)$ can also be computed efficiently using eigendecompositions of $L$ and related matrices (Borodin, 2009).

## 3 DPPs VS. MRFs

Given the tractability of DPPs, a natural question is how the representational power of DPPs compares to that of MRFs with repulsive potentials, which are generally intractable. Let $y_i = \mathbb{I}(i \in Y)$ be a binary indicator variable which is 1 if $i$ is selected and 0 otherwise. The MRF most closely resembling a DPP over $N$ items is a pairwise log-linear model over $y_1, \ldots, y_N$ with negative interaction terms:

$$P(\mathbf{Y} = Y) \propto \exp \left( \sum_i w_i y_i + \sum_{i<j} w_{ij} y_i y_j \right), \qquad (11)$$

where $w_{ij} \leq 0$. This model has the same number of parameters as the DPP, and can also capture negative correlations between variables $y_i$. The key representational difference is that, while $w_{ij}$ are individually constrained to be less than 0, the positive semi-definite constraint on the DPP kernel $L$ is global. In particular, similarities $S_{ij}$ must satisfy a kind of transitivity (if $S_{ij}$ and $S_{jk}$ are large, than so is $S_{ik}$), which is a consequence of the triangle inequality:

$$\sqrt{1 - S_{ij}^2} + \sqrt{1 - S_{jk}^2} \geq \sqrt{1 - S_{ik}^2} \ . \qquad (12)$$

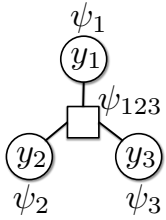

| $y_1y_2y_3$ | $\psi_{123}$-MRF | $\psi_{123}$-DPP |
|---|---|---|
| 000 | 1 | 1 |
| 001 | 1 | 1 |
| 010 | 1 | 1 |
| 100 | 1 | 1 |
| 110 | $e^{w_{12}}$ | $1 - S_{12}^2$ |
| 101 | $e^{w_{13}}$ | $1 - S_{13}^2$ |
| 011 | $e^{w_{23}}$ | $1 - S_{23}^2$ |
| 111 | $e^{w_{12}+w_{13}+w_{23}}$ | $1+2S_{12}S_{13}S_{23}$ $-S_{12}^2-S_{13}^2-S_{23}^2$ |

Figure 3: A factor graph representation of a 3-node MRF or DPP. The ternary factor $\psi_{123}$ is shown.

When $N = 2$, it is easy to show that the two models are equivalent, in the sense that they can both represent any distribution with negative correlations:

$$P(y_1 = 1)P(y_2 = 1) \geq P(y_1 = 1, y_2 = 1) \,. \quad (13)$$

When $N = 3$, the differences become apparent. In this setting there are six parameters for both the MRF $(w_1, w_2, w_3, w_{12}, w_{13}, w_{23})$ and the DPP $(q_1, q_2, q_3, S_{12}, S_{13}, S_{23})$. To place the two models on equal footing, we represent each as a factor graph with unnormalized node potentials $\psi_1, \psi_2, \psi_3$ and a single unnormalized ternary factor $\psi_{123}$ (see Figure 3). The node potentials are $\psi_i(y_i) = e^{w_i y_i}$ for the MRF and $\psi_i(y_i) = q_i^{2y_i}$ for the DPP. The ternary factors are:

$$\psi_{123}(y_1, y_2, y_3) = \exp\left(\sum_{i<j} w_{ij} y_i y_j\right) \quad \text{(MRF)}$$

$$\psi_{123}(y_1, y_2, y_3) = \det(S_Y) \quad \text{(DPP)}$$

Figure 3 shows the values of the ternary factors for all configurations. The last four entries are determined, respectively, by the three edge parameters of the MRF and three parameters $S_{ij}$ of the DPP, so the sets of realizable ternary factors form 3-D manifolds. We attempt to visualize these manifolds by showing 2-D slices for various values of the last entry (111) in Figure 4. To improve visibility, we have constrained $S_{12}, S_{13}, S_{23} \geq 0$ in the second and third panels. When this constraint is removed, the DPP surface is closed; the missing surface covers the triangular hole and bulges slightly away from the origin.

When entry 111 is large, negative correlations are weak and the two models give rise to qualitatively similar distributions. As the value of the entry 111 shrinks to zero, however, the surfaces become quite different. MRFs, for example, can describe a distribution where the first item is strongly anti-correlated with both of the others, but the second and third are not anti-correlated with each other. The transitive nature of the DPP makes this impossible. The third panel shows the MRF and DPP overlaid when the entry 111 is 0.1; even with these relatively strong interactions, the models are fairly close. We conjecture that the story for larger $N$ follows a similar pattern.

## 4 CONDITIONAL DPPs

For the purposes of modeling real data through discriminative learning, we propose a parameterized conditional DPP. For concreteness, consider extractive document summarization. In this setting, the input $X$ is a cluster of documents on a given topic, and the set $\mathcal{Y}(X)$ is a collection of candidate sentences from those documents. The goal is to choose a subset $Y \subseteq \mathcal{Y}(X)$, usually with some constraint on the total length of the sentences in $Y$, that best summarizes the input.

More generally, suppose we have an arbitrary input $X$ and set $\mathcal{Y}(X)$ of items derived from the input. A conditional DPP $\mathcal{P}(\mathbf{Y} = Y|X)$ is a conditional probabilistic model which assigns a probability to every possible subset $Y \subseteq \mathcal{Y}(X)$. Our model takes the form of an L-ensemble:

$$\mathcal{P}(\mathbf{Y} = Y|X) \propto \det(L_Y(X)) \,, \quad (14)$$

where $L(X)$ is a positive semidefinite $|\mathcal{Y}(X)| \times |\mathcal{Y}(X)|$ kernel matrix that depends on the input.

As discussed in Section 2, the normalization can be computed efficiently and is given by $\det(I + L(X))$. Using the $q/\phi$ decomposition given in Equation (8), we have

$$L_{ij}(X) = q_i(X)\phi_i(X)^\top \phi_j(X)q_j(X) \quad (15)$$

for suitable $q_i(X) \in \mathbb{R}^+$ and $\phi_i(X) \in \mathbb{R}^n$, $\|\phi_i(X)\|_2 = 1$. While similarity features $\phi_i(X)$ can consist of any desired measurements (and can even be infinite-dimensional, as long as the resulting similarity matrix $S$ is a proper kernel), we compute the quality scores using a log-linear model:

$$q_i(X) = \exp\left(\frac{1}{2}\theta^\top \boldsymbol{f}_i(X)\right) \,, \quad (16)$$

where $\boldsymbol{f}_i(X) \in \mathbb{R}^m$ is a feature vector for item $i$ and $\theta \in \mathbb{R}^m$ is a parameter vector. Note that feature vectors $\boldsymbol{f}_i(X)$ are in general distinct from $\phi_i(X)$; the former are used for modeling quality, while the latter measure similarity.

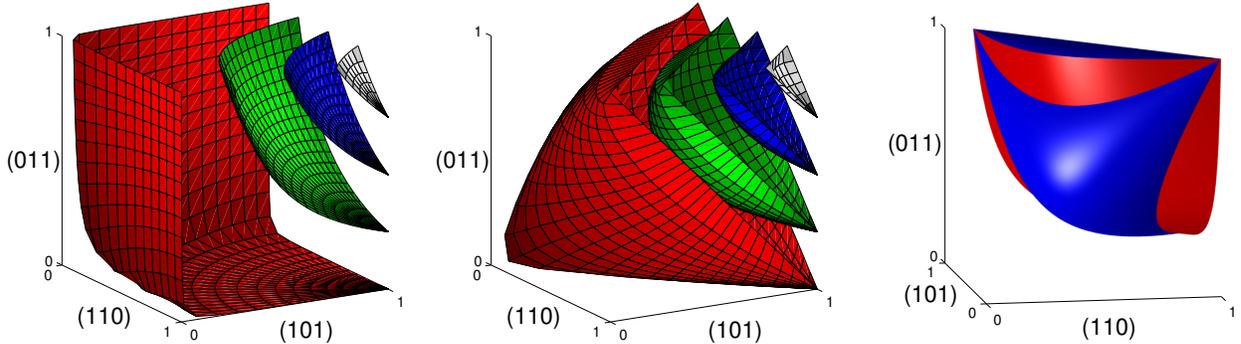

Figure 4: Realizable values of entries 110, 101, 011 in a 3-factor when 111 = 0.001 (red), 0.25 (green), 0.5 (blue), and 0.75 (grey). **Left:** MRF. **Middle:** DPP. **Right:** Superimposed MRF (red) and DPP (blue) surfaces when 111 = 0.1 (figure rotated for better perspective).

## 5 LEARNING

Assume now that we are given a training set consisting of $T$ pairs $(X^1, Y^1), \ldots, (X^T, Y^T)$, where each $X^t$ is an input (e.g., a document cluster) and the corresponding $Y^t \in \mathcal{Y}(X^t)$ is a selected subset (e.g., an "ideal" extractive summary). To learn the parameters $\theta$ of our model, we maximize the log-likelihood of the training set:

$$\mathcal{L}(\theta) = \log \prod_t \mathcal{P}_\theta(Y^t|X^t) = \sum_t \log \mathcal{P}_\theta(Y^t|X^t) \ . \quad (17)$$

For ease of notation, going forward we assume that the training set contains only a single instance and drop the instance index $t$. Of course, this does not change the generality of the results.

**Proposition 1.** $\mathcal{L}(\theta)$ is concave in $\theta$.

*Proof.*

$$\log \mathcal{P}_\theta(Y|X) = \theta^\top \sum_{i \in Y} \boldsymbol{f}_i(X) + \log \det(S_Y(X))$$

$$- \log \sum_{Y'} \exp\left(\theta^\top \sum_{i \in Y'} \boldsymbol{f}_i(X)\right) \det(S_{Y'}(X)) \ . \quad (18)$$

With respect to $\theta$, the first term is linear, the second is constant, and the third is the composition of a concave function (negative log-sum-exp) and a non-negative linear function, so the expression is concave. □

We can therefore apply standard convex optimization techniques as long as we can efficiently compute the gradient $\nabla\mathcal{L}(\theta)$. As in standard maximum entropy modeling, the gradient of the log-likelihood can be seen as the difference between the empirical feature counts and the expected feature counts under the model distribution:

$$\nabla\mathcal{L}(\theta) = \sum_{i \in Y} \boldsymbol{f}_i(X) - \sum_{Y'} \mathcal{P}_\theta(Y'|X) \sum_{i \in Y'} \boldsymbol{f}_i(X) \ . \quad (19)$$

The sum over $Y'$ is exponential in $|\mathcal{Y}(X)|$; hence we cannot compute it directly. Instead, note that we can rewrite:

$$\sum_{Y'} \mathcal{P}_\theta(Y'|X) \sum_{i \in Y'} \boldsymbol{f}_i(X) = \sum_i \boldsymbol{f}_i(X) \sum_{Y' \supseteq \{i\}} \mathcal{P}_\theta(Y'|X).$$

---

**input** : instance $(X, Y)$, parameters $\theta$
**output**: gradient $\nabla\mathcal{L}(\theta)$

1 Compute $L(X)$ as in Equation (8);
2 Eigendecompose $L(X) = \sum_k \lambda_k v_k v_k^\top$;
3 **for** $i \in \mathcal{Y}(X)$ **do**
4 $\quad K_{ii} \leftarrow \sum_k \frac{\lambda_k}{1+\lambda_k} v_{ki}^2$;
5 **end**
6 $\nabla\mathcal{L}(\theta) \leftarrow \sum_{i \in Y} \boldsymbol{f}_i(X) - \sum_i K_{ii} \boldsymbol{f}_i(X)$;

**Algorithm 1:** Gradient of the log-likelihood.

---

That is, we can compute the expected feature counts if we know the marginal probability of inclusion for each sentence $i$. Recall that for DPPs, these probabilities are given by the diagonal of $K$, the marginal kernel, which can be computed through an eigendecomposition of the kernel $L(X)$. Thus, we can efficiently compute the gradient as described in Algorithm 1.

## 6 MAP INFERENCE

At test time, we need to take the learned parameters $\theta$ and use them to predict a set $Y$ for a previously unseen input $X$. One option is to sample from the conditional distribution, which can be done exactly for DPPs in cubic time (Hough et al., 2006; Kulesza and Taskar, 2010). However, we found that we obtained better performance with the maximum *a posteriori* (MAP) set, conditioned on a budget constraint:

$$Y^{MAP} = \arg\max_{Y} \mathcal{P}_\theta(Y|X)$$
$$\text{s.t.} \sum_{i \in Y} \text{cost}(i) \leq B \ , \qquad (20)$$

where $\text{cost}(i)$ is some measure of the cost of item $i$, and $B$ is a limit on the total cost. Computing $Y^{MAP}$ exactly is NP-hard (Ko et al., 1995), but we consider two approximations. First, we can obtain a brute-force estimate by sampling a large number of sets $Y$, and then choosing the sample that has highest probability among all those satisfying the budget constraint. Second, and more practically, we note that the optimization in Equation (20) is submodular, and so we can approximate it through a simple greedy algorithm.

---

**input** : input $X$, parameters $\theta$, budget $B$
**output**: set $Y$
**1** $U \leftarrow \mathcal{Y}(X); Y \leftarrow \emptyset$;
**2 while** $U \neq \emptyset$ **do**
**3** $\quad Y \leftarrow Y \cup \arg\max_{i \in U} \left( \frac{\mathcal{P}_\theta(Y \cup \{i\}|X) - \mathcal{P}_\theta(Y|X)}{\text{cost}(i)} \right)$;
**4** $\quad U \leftarrow U \setminus \{i \mid \text{cost}(Y) + \text{cost}(i) > B\}$;
**5 end**
**6** return $Y$;

**Algorithm 2:** Approximately computing the MAP set under a budget constraint.

---

This algorithm is closely related to those given in Lin and Bilmes (2010) and Krause and Guestrin (2005). For monotone submodular problems, algorithms of this type have formal approximation guarantees. Our MAP problem is not generally monotone; nonetheless, Algorithm 2 seems to work well in practice.

## 7 SUMMARIZATION

We employ learning of conditional DPPs to perform extractive summarization of text. As in Section 4, the input $X$ is a cluster of documents, and $\mathcal{Y}(X)$ is a set of candidate sentences from those documents. In our experiments $\mathcal{Y}(X)$ contains all sentences in the cluster, though preprocessing can also be used to try to improve the candidate set (Conroy et al., 2004). Given a predicted set of sentences $Y$, we follow previous work and construct a final summary by placing the sentences in the same order in which they appeared in the original documents (Lin and Bilmes, 2010). This may not be optimal; however, it is partially justified by the fact that modern automatic evaluation metrics, such as ROUGE, are mostly invariant to sentence order.

Following previous work, we experiment with data from the Document Understanding Conference (DUC) 2003 and 2004 multi-document summarization task (Task 2) (Lin and Bilmes, 2010; Dang, 2005). The task is to generate event-focused summaries of document clusters from the NIST TDT collection. Each cluster contains approximately 10 articles from the AP and New York Times newswires, and covers a single topic over a short time span. The clusters have a mean length of approximately 250 sentences and 5800 words. The 2003 task, which we use for training, contains 30 clusters, and the 2004 task, which is our test set, contains 50 clusters. Summaries are required to be at most 665 bytes in length, including spaces; this forms our budget constraint. Each cluster comes with four reference human summaries for evaluation purposes. Figure 5 depicts a sample cluster from the test set.

To measure performance we use ROUGE, an automatic evaluation metric for summarization (Lin, 2004). ROUGE measures n-gram overlap statistics between the human references and the summary being scored, and combines them to produce various sub-metrics. ROUGE-1, which is a simple unigram recall measure, has been shown to correlate quite well with human judgments (Lin, 2004). We therefore follow previous work and use ROUGE's unigram F-measure as our primary metric. We refer to this measure as ROUGE-1F. We also report ROUGE-1P and ROUGE-1R (precision and recall, respectively) as well as ROUGE-2F and ROUGE-SU4F, which include bigram match statistics and have also been shown to correlate well with human judgments. Our implementation uses ROUGE version 1.5.5 with stemming turned on, but without stopword removal. These settings correspond to those used for the actual DUC competitions (Dang, 2005).

### 7.1 BUILDING A TRAINING SET

The human reference summaries provided with the DUC data are high-quality, but to train our model we need target *extractive* summaries. To obtain high-quality extractive "oracle" summaries for training, we employ a simple greedy algorithm. Intuitively, the idea is to choose at each round the sentence that achieves maximal mean unigram F-measure to the human references, and add it to the summary. Since high F-measure requires high precision as well as recall, we then update the references by removing the words "covered" by the newly selected sentence, and proceed to the next round.

Applying this algorithm to the DUC 2003 training data, we achieve the ROUGE scores shown in Table 1. We compare the scores of our oracle extractive summaries to those of the best automatic system from the DUC competition in 2003, as well as the actual human references. Note that, in the third column, the human summary being evaluated is also one of the four references used to compute ROUGE; hence the scores are probably significantly higher than could be achieved in practice. Furthermore, it has been shown that extractive summaries, even when generated optimally, are by nature limited in quality compared to unconstrained summaries (Genest et al., 2010). Thus we believe that our oracle summaries are excellent targets for training.

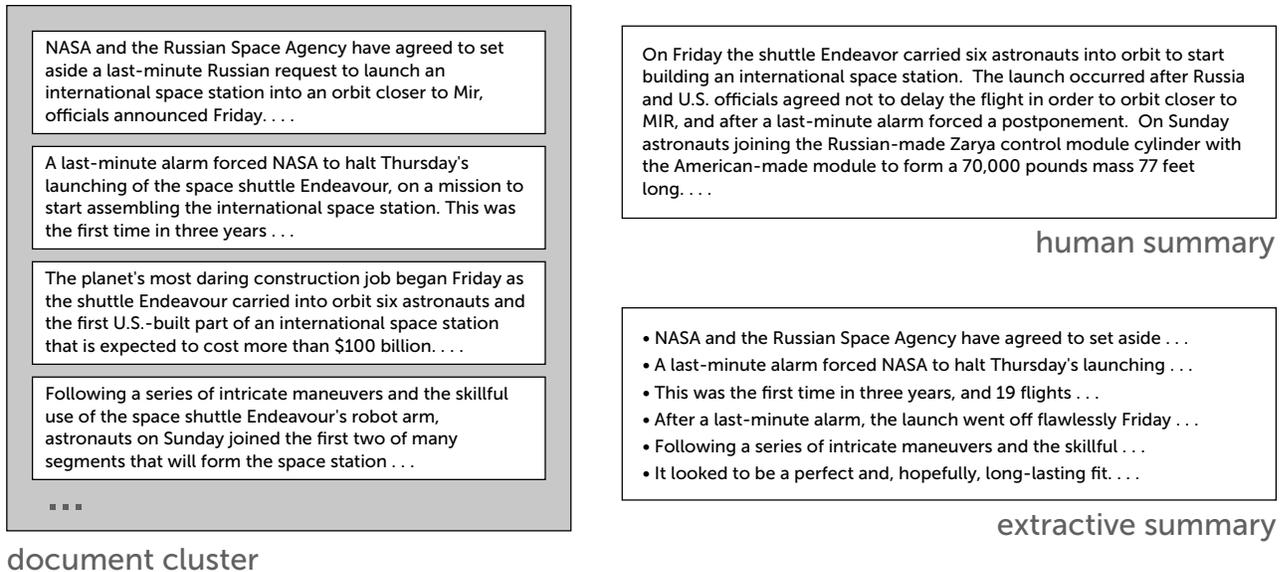

Figure 5: A sample cluster from the DUC 2004 test set, with one of the four human reference summaries and a (manufactured) extractive summary.

Table 1: ROUGE scores for the best automatic system from DUC 2003, our heuristically-generated oracle extractive summaries, and human summaries.

| System | ROUGE-1F | ROUGE-2F | ROUGE-SU4F |
|---|---|---|---|
| Machine | 35.17 | 9.15 | 12.47 |
| Oracle | 46.59 | 16.18 | 19.52 |
| Human | 56.22 | 33.37 | 36.50 |

### 7.2 FEATURES

For similarity features $\phi_i(X)$, we use standard normalized tf-idf vectors, where the idf scores are computed across all of the documents in all of the clusters in the training set. Thus the similarity between two sentences is just their cosine distance. We do not use stop word removal or stemming. We augment the tf-idf vectors with an additional constant feature taking the value $\rho$. This has the effect of making all sentences more similar to one another, increasing repulsion. We set $\rho$ to optimize ROUGE-1F score on the training set; in our experiments, the best choice was $\rho = 0.3$.

We have the following quality features $\boldsymbol{f}_i(X)$. For some of the features, we make use of cosine distances; these are computed using the same tf-idf vectors as the similarity features. When a feature is intrinsically real-valued, we produce a series of binary features by binning. The bin boundaries are determined either globally or locally. Global bins are evenly spaced quantiles of the feature values across all sentences in the training set, while local bins are quantiles of the feature values in the current cluster. Below, we indicate which types of bins are used in each case.

- **Constant**: A constant feature allows the model to bias towards summaries with more or fewer sentences.

- **Length**: We bin the length of the sentence (in bytes) into five global bins.

- **Document position**: We compute the position of the sentence in its original document and generate binary features indicating positions 1–5, plus a sixth binary feature indicating all other positions. We expect that, for newswire text, sentences that appear earlier in an article are more likely to be useful for summarization.

- **Mean cluster similarity**: For each sentence we compute the average cosine distance to all other

sentences in the cluster. This attempts to measure how well the sentence reflects the salient words occurring most frequently in the cluster. We use the raw score, five global bins, and ten local bins.

- **LexRank**: We compute continuous LexRank scores by finding the principal eigenvector of the row-normalized cosine similarity matrix. (See Erkan and Radev (2004) for details.) This provides an alternative measure of centrality. We use the raw score, five global bins, and five local bins.

In total we have 39 quality features; including $\rho$ our model has 40 parameters.

### 7.3 RESULTS

We train our model with a standard L-BFGS optimization algorithm. We place a zero-mean Gaussian prior on the parameters $\theta$, with variance set to optimize ROUGE-1F on a development subset of the 2003 data. We learn parameters $\theta$ on the DUC 2003 corpus, and test them using DUC 2004 data.

We test our model with the two MAP inference algorithms described in Section 6. The first is a brute-force estimate of the optimal summary, obtained by taking $10^6$ samples from the model, then choosing the highest-probability sample with length between 660 and 680 characters. (Overlong summaries are trimmed to the limit of 665 bytes.) This process is computationally expensive, but it provides a point of comparison for the much more efficient greedy procedure.

As a simple baseline, we compare against the first 665 bytes of the cluster text. Since the clusters consist of news articles, this is not an unreasonable summary in many cases. We refer to this baseline as Begin.

We test two additional baselines that rely on quality scores trained by logistic regression, where each sentence is a unique instance to be classified as included or not included, using labels from our training oracle and the features described above. For baseline LR+MMR, the quality scores are used for the relevance portion of the well-known maximum marginal relevance (MMR) algorithm (Carbonell and Goldstein, 1998). Sentences are chosen iteratively according to

$$\underset{i \in \mathcal{Y}(X)}{\arg\max} \left[ \lambda q_i(X) - (1-\lambda) \max_{j \in Y} S_{ij} \right] , \quad (21)$$

where $Y$ is the set of sentences already selected (initially empty), $q_i(X)$ is the learned quality score, and $S_{ij}$ is the cosine similarity between sentences $i$ and $j$. The tradeoff $\lambda$ is optimized on a development set, and sentences are added until the budget is full. For baseline LR+DPP, the logistic regression quality scores are used directly in our DPP model, with inference performed using Algorithm 2; comparing to this baseline allows us to isolate the contribution of learning the model parameters in context.

For more sophisticated baselines, we compare against the three highest-scoring systems from DUC 2004—peers 65, 104, and 35—as well as the submodular graph-based approach recently described by Lin and Bilmes (2010), which we refer to as SubMod. Because code was not available for the latter system, we implemented it ourselves, striving to replicate their results. Our version fails to reproduce the exact numbers given in prior work; however, we suspect this is due to differences in the text processing pipeline, including detagging, sentence splitting, and the computation of tf-idf scores. Because the DPP model relies upon the same tf-idf scores as SubMod, any change to the preprocessing is likely to impact the performance of both systems. We set the hyperparameters for SubMod to the values reported by Lin et al ($r = 0.3, \lambda = 4$).

Table 2 shows the results for all methods on the DUC 2004 test corpus. Scores for the actual DUC competitors differ slightly from the originally reported results because we use an updated version of the ROUGE package.

## 8 CONCLUSION

We showed how determinantal point processes can be applied to subset selection tasks like extractive summarization, where negative interactions are key and traditional models like Markov random fields can be computationally problematic. DPPs give rise to an intuitive tradeoff between high-quality items, like sentences that concisely represent the text to be summarized, and diverse items, like sentences that reduce redundancy and increase coverage. We showed how to parameterize the model to allow for convex training, efficiently maximizing the likelihood of a training set despite the exponential number of possible outputs. We also showed how to perform efficient approximate test-time inference. Experiments on the DUC 2004 multi-document summarization task demonstrate state-of-the-art performance.

We believe that learning DPPs has many applications beyond summarization. For example, diversity is an important consideration in many search applications: we would like to return results that are not only relevant to the user, but also cover multiple possible interpretations of ambiguous queries. A search for "apple", for example, might be a request for information about fruit, computers, record companies, etc.; if we don't know which one the user is after, it might be preferable to return a few results on each topic, rather than a full set of results about whichever happens to be most common.

Table 2: ROUGE scores on the DUC 2004 test set.

| System | ROUGE-1F | ROUGE-1P | ROUGE-1R | ROUGE-2F | ROUGE-SU4F |
|---|---|---|---|---|---|
| Begin | 32.08 | 31.53 | 32.69 | 6.52 | 10.37 |
| LR+MMR | 36.01 | 35.36 | 36.73 | 7.46 | 11.87 |
| LR+DPP | 37.48 | 37.18 | 37.82 | 8.54 | 12.74 |
| Peer 35 | 37.54 | 37.69 | 37.45 | 8.37 | 12.90 |
| Peer 104 | 37.12 | 36.79 | 37.48 | 8.49 | 12.81 |
| Peer 65 | 37.87 | 37.58 | 38.20 | 9.13 | 13.19 |
| SubMod | 37.39 | 36.86 | 37.99 | 8.42 | 12.78 |
| DPP | **38.27** | 37.87 | **38.71** | **9.35** | **13.45** |
| DPP (greedy) | 38.20 | **37.97** | 38.46 | 9.14 | 13.26 |

Future work on summarization with DPPs includes combining the extractive approach with internal sentence compression to achieve more informative summaries (Martins and Smith, 2009; Clarke and Lapata, 2008). This approach requires treating each sentence as a structure rather than a simple binary in-or-out decision. Recent work has shown that DPPs can be extended to efficiently deal with such structures in many cases (Kulesza and Taskar, 2010).

More generally, an important open question regarding DPPs is whether it might be possible to also parameterize the similarity matrix in a way that permits efficient learning. This would enable a wider variety of more expressive models, analogous to the advantages offered by kernel matrix learning for SVMs. Additionally, while we illustrated the relationship between MRFs and DPPs for $N = 3$, extending these observations to larger $N$ might lead to improved inference algorithms for a subclass of "DPP-like" MRFs; e.g., using variational methods.

**Acknowledgments**

The authors were partially supported by NSF Grant 0803256.